%% file: main.tex
\documentclass[11pt]{article}

\usepackage[preprint]{acl}

\usepackage{times}
\usepackage{latexsym}
\usepackage{amsmath}
\usepackage{amssymb}
\usepackage[T1]{fontenc}

\usepackage[utf8]{inputenc}

\usepackage{microtype}

\usepackage{inconsolata}

\usepackage{graphicx}

\usepackage{booktabs}
\usepackage{float}

\newcommand{\harvard}{%
    \includegraphics[width=0.02\textwidth]{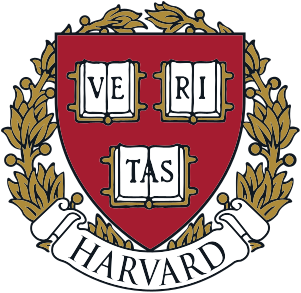} 
}
\newcommand{\stanford}{%
    \includegraphics[width=0.014\textwidth]{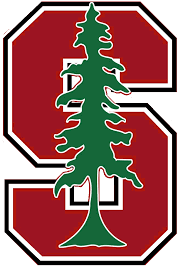} 
}
\newcommand{\cmu}{%
    \includegraphics[width=0.016\textwidth]{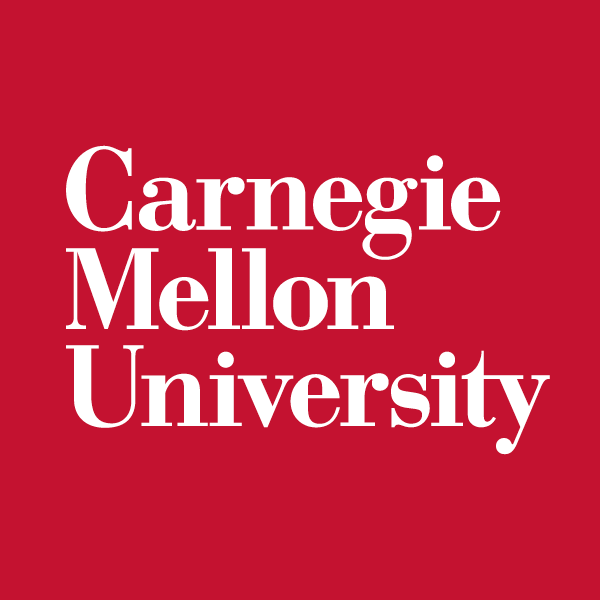} 
}

\title{LLM Review: Enhancing Creative Writing via Blind Peer Review Feedback}

\author{Weiyue Li\thanks{Equal contribution}\harvard \quad
Mingxiao Song$^{*}$\harvard \quad
Zhenda Shen$^{*}$\harvard \quad
Dachuan Zhao$^{*}$\harvard \quad \\
\textbf{Yunfan Long}\cmu \quad
\textbf{Yi Li}\cmu \quad
\textbf{Yongce Li}\stanford \quad
\textbf{Ruyi Yang}\harvard \quad
\textbf{Mengyu Wang}\harvard \\
\harvard Harvard University, \cmu Carnegie Mellon University, \stanford Stanford University
}

\newcommand{\llmreview}{\textit{LLM Review}}

\usepackage[dvipsnames]{xcolor}
\usepackage{tcolorbox}

\definecolor{sciintcol}{HTML}{E6F0FF} 
\definecolor{speclogcol}{HTML}{E9F8EF} 
\definecolor{chardepcol}{HTML}{F3E8FF} 
\definecolor{worldcol}{HTML}{FFF3E0}   
\definecolor{ethicscol}{HTML}{FFF0F5}  

\newcommand{\sciintaspect}[1]{%
  \colorbox{sciintcol}{\parbox{\dimexpr\linewidth-2\fboxsep\relax}{#1}}%
}
\newcommand{\speclogaspect}[1]{%
  \colorbox{speclogcol}{\parbox{\dimexpr\linewidth-2\fboxsep\relax}{#1}}%
}
\newcommand{\chardepaspect}[1]{%
  \colorbox{chardepcol}{\parbox{\dimexpr\linewidth-2\fboxsep\relax}{#1}}%
}
\newcommand{\worldaspect}[1]{%
  \colorbox{worldcol}{\parbox{\dimexpr\linewidth-2\fboxsep\relax}{#1}}%
}
\newcommand{\ethicsaspect}[1]{%
  \colorbox{ethicscol}{\parbox{\dimexpr\linewidth-2\fboxsep\relax}{#1}}%
}

\begin{document}
\maketitle

\begin{abstract}

Large Language Models (LLMs) often struggle with creative generation, and multi-agent frameworks that improve reasoning through interaction can paradoxically hinder creativity by inducing content homogenization. We introduce LLM Review, a peer-review-inspired framework implementing Blind Peer Review: agents exchange targeted feedback while revising independently, preserving divergent creative trajectories. To enable rigorous evaluation, we propose SciFi-100, a science fiction writing dataset with a unified framework combining LLM-as-a-judge scoring, human annotation, and rule-based novelty metrics. Experiments demonstrate that LLM Review consistently outperforms multi-agent baselines, and smaller models with our framework can surpass larger single-agent models, suggesting interaction structure may substitute for model scale.\footnote{Our \href{https://github.com/weiyueli7/llm-review}{code and data}}

\end{abstract}

\input{Section/1intro}
\input{Section/2related}
\input{Section/3method}

\input{Section/4exp}

\input{Section/5results}
\input{Section/6dis}

\section*{Limitations}

Our evaluation focuses on short-form science fiction writing; generalization to other creative domains (poetry, long-form fiction, music) may require domain-specific metrics and reference corpora. Our rule-based novelty metrics measure divergence from a fixed reference corpus (SFGram) and do not by themselves guarantee meaningful creativity; we therefore interpret them jointly with quality-oriented LLM-as-a-judge scores. Our human study uses nine student annotators on a single configuration; professional writers might assess differently. Finally, LLM Review requires approximately 9× the inference cost of single-agent generation, though this can be offset by using smaller models.

\section*{Ethical consideration}
This work includes human evaluation of machine-generated text; annotators provided informed consent, no personal identifying information was collected, and results are reported in aggregate. Generated content may reflect biases present in underlying language models, particularly in speculative narratives, and automated critique could reinforce shared assumptions. The proposed framework is intended for research use and should be deployed with human oversight in practical applications.

\bibliography{custom}

\appendix

\input{Section/7appendix}

\end{document}

%% file: Section/1intro.tex
\section{Introduction}
\label{sec:intro}

Large Language Models (LLMs) have achieved strong performance across natural language processing tasks \cite{lappin2024assessing, jin2024comprehensive, yi2024survey}, and are increasingly deployed in multi-agent systems for reasoning and coordination \cite{tran2025multiagentcollaborationmechanismssurvey}. However, these systems are optimized for correctness rather than creativity. Prior work finds that LLMs tend to reproduce familiar patterns rather than generate genuinely novel ideas \cite{mohammadi2024creativity, chakrabarty2024art, li2025automated}, and existing single-agent approaches such as decoding strategies, prompt engineering, and post-training optimization \cite{yang2024largelanguagemodelsoptimizers, potraghloo2025tophdecodingadaptingcreativity, matan2025comprehensive} yield surface-level diversity rather than substantive conceptual novelty.

Human creativity is fundamentally social, emerging through discussion, critique, and iterative refinement \cite{paulus2003group}. Recent multi-agent frameworks such as debate and discussion show improvements in reasoning and output diversity \cite{Du2023ImprovingFA, lu2024llmdiscussionenhancingcreativity, summers2023brainstorm}, sharing an implicit assumption: more interaction yields better outcomes. We argue this assumption breaks down for creativity. Research on group brainstorming shows that interactive groups often produce fewer and less original ideas than individuals working independently, due to production blocking and convergent tendencies \cite{diehl1987productivity, larey1999group}. Recent work further demonstrates homogenization effects when humans collaborate with LLMs \cite{anderson2024homogenization}. We propose a different view: creativity is not improved by more interaction, but by the right information flow constraints. Creative novelty requires divergence, the ability to explore different trajectories without converging on share patterns \cite{gillebaart2013unraveling}. Existing frameworks repeatedly expose agents to each other's evolving outputs, inadvertently encouraging alignment and limiting semantic exploration.

We introduce \textbf{LLM Review}, a framework that enhances creativity by constraining rather than maximizing information flow through a mechanism we call Blind Peer Review. Inspired by double-blind academic reviewing, agents provide targeted feedback on peers' initial drafts but revise independently, without seeing how peers respond to the same feedback. This information asymmetry lets agents benefit from external critique while preserving independent creative trajectories. To evaluate our approach, we introduce \textbf{SciFi-100}, a science fiction writing dataset, together with a unified evaluation framework combining LLM-as-a-judge scoring, human annotation, and rule-based metrics capturing lexical and semantic novelty against a corpus of canonical science fiction \cite{colton2008creativity}.

Our contributions: (1) \textbf{LLM Review}, a framework that enhances creativity by constraining information flow through Blind Peer Review; (2) \textbf{SciFi-100}, the first science fiction writing dataset with a unified evaluation framework combining LLM-as-a-judge, human annotation, and rule-based novelty metrics; (3) LLM Review outperforms baselines, with smaller models exceeding larger single-agent models, showing that interaction structure can offset model scale.

%% file: Section/2related.tex
\section{Related Work}
\label{sec:related}

\paragraph{Multi-Agent LLMs}
Recent work has explored multi-agent frameworks built on Large Language Models (LLMs) to improve factuality, reasoning, and task performance through structured interaction, including role-based workflows, debate and critique protocols, persona-driven interaction, and large-scale orchestration \cite{yao2022react,chen2023reconcile,hong2024metagpt,qian-etal-2024-chatdev,Du2023ImprovingFA,chan2023chatevalbetterllmbasedevaluators,tseng2024talespersonallmssurvey,wang-etal-2025-megaagent}. These systems are primarily evaluated on goal-directed benchmarks and focus on task success, autonomy, or efficiency \cite{li2025agent_oriented_planning,ye2025mas_gpt,qian2025scaling_macnet,dang2025evolving_orchestration,zhang2025cut}. In contrast, our work studies creative writing and shows that interaction \emph{structure}, rather than interaction frequency or scale, plays a critical role: strategically restricting information flow helps preserve divergent creative trajectories.

\paragraph{LLM Creativity}
Creativity in language models has been studied across tasks such as literary composition, metaphor generation, and alternative use generation, primarily in single-LLM settings \cite{gómezrodríguez2023confederacymodelscomprehensiveevaluation,chakrabarty2024art,doi:10.1080/10400419.2024.2326343,stevenson2022puttinggpt3screativityalternative}. Prior approaches improve creativity through decoding strategies, post-training optimization, or inference-time prompting \cite{ghazvininejad2017hafez,keskar2019ctrl,wei2025ignitingcreativewritingsmall,chung2025modifyinglargelanguagemodel,lagzian2025multinoveltyimprovediversitynovelty}. While effective, these methods largely operate at model or decoding level; our work instead frames creativity as a socially grounded, multi-agent process driven by structured discussion.

\paragraph{Creativity Evaluation}
Evaluating creativity is inherently subjective, and prior work relies on human judgments or LLM-based evaluators for scalability \cite{gómezrodríguez2023confederacymodelscomprehensiveevaluation,chakrabarty2024art,feng2024sampleefficienthumanevaluationlarge,zheng2023judging}. Automatic proxies are often used to capture complementary aspects such as diversity and semantic novelty relative to a reference corpus \cite{zhang2020tradingdiversityqualitynatural,peeperkorn2024temperaturecreativityparameterlarge}. Following established views of creativity as balancing novelty and value \cite{colton2008creativity,d2021characterises}, we adopt a combined evaluation framework that integrates rule-based novelty metrics with LLM-as-a-judge assessment.

%% file: Section/3method.tex
\section{Methodology}
\label{sec:method}

\subsection{Task Definition}
\label{sec:task}
Given a science-fiction writing prompt $x \in \mathcal{X}$, the goal is to generate a short story $y$ (approximately 300 words) that is coherent, creatively strong, and of good quality. We study both single-agent generation and multi-agent discussion frameworks that iteratively improve drafts through structured interaction. Unless otherwise specified, all frameworks use the same base writer model (the model under evaluation) and the same role-played writer personas to control for style and diversity effects.

\subsection{SciFi-100 Data Curation}
To curate high-quality science fiction prompts that can be used to evaluate model performance, we structure a systematic process to create a dataset that matches creative writing attributes. We first identify ten central aspects (see Appendix~\ref{apen:scifi-100}) of creative writing based on our experiences as human writers as well as foundational insights from narratology, literary theory, and creative writing pedagogy~\citep{genette1980narrative, pound2013abc, freytag1895technique, forster1927aspects,rosenblatt1994reader,burroway2022writing,csikszentmihalyi1990domain,chatman1978story,ricoeur2004rule,tannen2005conversational}. For each aspect, we query LLMs~\citep{openai2024gpt4ocard} to generate twenty unique prompts by specifically instructing the model to create scenarios where a science fiction narrative can unfold. After the model generates 200 prompts (20 prompts per writing aspect), we manually select and revise 10 prompts per aspect (100 in total) to ensure diversity and thematic relevance. The dataset’s balanced distribution across aspects ensures a comprehensive evaluation of creative dimensions. To our best knowledge, SciFi-100 is the first dataset designed to assess scientific fiction writings of LLMs.  

\subsection{Multi-Agent Role-Play Setup}
\label{sec:roleplay_setup}
For multi-agent frameworks, we instantiate $N{=}3$ writer agents. Following prior work on role-play and diversity of thought~\citep{design_thinking, lu2024llmdiscussionenhancingcreativity}, each agent is assigned a persistent persona (e.g., Humanistic Writer, Futuristic Writer, Ecological Writer) that remains fixed across all rounds. We repeatedly restate these roles in prompts to encourage consistent viewpoints and reduce homogenization. All frameworks share the same formatting constraints (story-only outputs, no commentary) to minimize evaluation noise.

\subsection{Compared Frameworks}
\label{sec:frameworks}
We compare our proposed framework, LLM Review, against the following baselines.

\paragraph{Single Agent} A single LLM is prompted once to write a $\sim$300-word science-fiction story for the given prompt. This baseline captures the base model's inherent creative writing capability under zero-shot prompting.

\paragraph{LLM Teacher}
LLM Teacher, inspired by classroom-style role-play prompting \citep{design_thinking}, models a teacher-student loop in which a teacher agent provides guidance and critique to student writers. The framework proceeds in three phases: the teacher offers high-level advice, students draft stories and receive aggregated feedback, and students revise to produce final outputs. This baseline represents a simple extension of role-play prompting with critique, but its teacher-centered, one-to-many feedback can encourage convergence toward similar revisions.

\paragraph{LLM Debate} LLM Debate~\citep{Du2023ImprovingFA} structures interaction as proposal and critique, where agents present candidate drafts and challenge each other’s content (logic gaps or weak originality), followed by refinement. Unlike other role-play-based frameworks, LLM Debate doesn't assign explicit personas to agents. We adapt the debate protocol for creative writing by focusing critiques on plausibility, novelty, and narrative quality.

\paragraph{LLM Discussion} LLM Discussion~\citep{lu2024llmdiscussionenhancingcreativity} is a three-phase multi-agent framework (Initiation, Discussion, Convergence) built on top of~\citet{Du2023ImprovingFA} with role-play. Agents iteratively read others' drafts and update their responses accordingly. This baseline represents structured multi-agent collaboration without explicit critique roles.

\begin{figure*}[!ht]
    \centering
\includegraphics[width=1\linewidth]{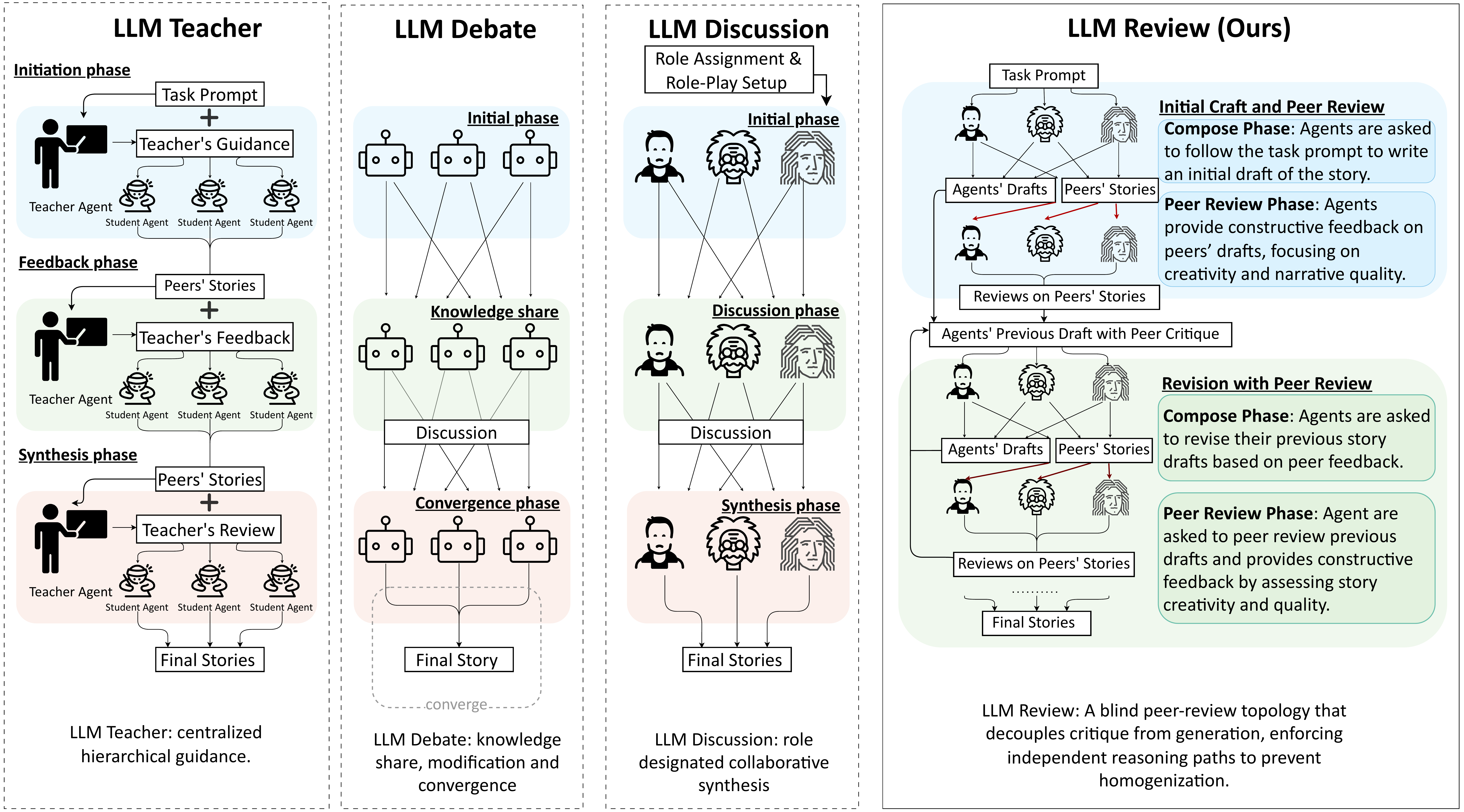}
  \caption{Comparison of multi-agent framework, from single-agent zero-shot writing to multi-agent frameworks. A single LLM generates a story in one pass without feedback, while LLM Teacher, LLM Debate, and LLM Discussion introduce hierarchical guidance, discussion, or role-based collaboration. LLM Review (ours) adopts a blind peer-review topology that decouples critique from generation, enabling independent revisions and reducing homogenization.}
  \label{fig:framework}
\end{figure*}

\subsection{LLM Review}
\label{sec:review_method}
Inspired by the iterative peer-review process in academic writing, we propose \llmreview, a structured feedback loop designed to improve creativity through \emph{distributed critique} and \emph{private revision}. The key idea is to let agents both \emph{create} and \emph{act as reviewers} for each other, then revise using feedback without seeing peers' revised drafts.

\paragraph{Phase 1: Compose}
Each agent independently writes an initial story draft conditioned only on the prompt and its persona.

\paragraph{Phase 2: Review}
Each agent reviews peers' drafts and provides targeted feedback (e.g., originality, world-building opportunities, speculative consistency, stronger imagery, character depth). Agents then revise their own draft using (a) their initial draft and (b) the received feedback.

\paragraph{Originality constraint}
During revision, agents do \textbf{not} see peers' revised drafts (only the initial drafts and feedback). This design reduces homogenization and preserves independent creative trajectories across rounds.

Figure~\ref{fig:framework} illustrates the interaction flow of LLM Review and the corresponding baselines.

\subsection{LLM-as-a-Judge Evaluation}
~\label{sec:llmasajudge}
LLMs have demonstrated the ability to approximate human judgment and effectively evaluate content, achieving results comparable to human evaluators, even on creativity tasks~\citep{lu2024llmdiscussionenhancingcreativity}. Therefore, we adopt an LLM-as-a-judge technique to assess the creativity of the generated stories, utilizing \texttt{gpt-4o}~\citep{openai2024gpt4ocard} as the judging model.

In our evaluation pipeline, each story is individually assessed on five key aspects derived from the literature on science fiction writing: \textbf{Scientific Concept Integration}, \textbf{Speculative Logic}, \textbf{Character Depth}, \textbf{Immersive World-Building}, and \textbf{Ethical and Philosophical Themes}. These evaluation aspects reflect the consensus of prior work in narratology, speculative fiction studies, creativity research, and narrative ethics~\citep{canavan2016metamorphoses, chatman1978story, forster1927aspects, le2015steering, nussbaum1988love}. Following~\citet{li2026grading}, for each aspect, the LLM assigns a score between 0 and 5, where 0 indicates potential plagiarism with poor quality, and 5 indicates highly creative writings of good quality. To ensure consistency and minimize variability, we evaluate each story three times and report the average score.

\subsection{Human evaluation}

To validate that our LLM-as-a-Judge scores reflect human preferences, we recruit nine student annotators to evaluate the stories generated on SciFi-100 by the LLM-Review framwork with Llama-3.2-3B model. Annotators rate each story using the same five criteria as in Section~\ref{sec:llmasajudge} and the same 0-5 Likert rubric. For each story and criterion, we average the nine ratings to obtain a single human consensus score. We report the human score distribution (mean $\pm$ std over stories) and in Table~\ref{tab:human}.

\subsection{Rule-based Evaluation}

To evaluate the creativity of generated content, we design a rule-based evaluation framework with two components. First, we measure an intrinsic dimension: the absolute token-level diversity of the generated text. Second, we measure novelty relative to traditional science fiction, and decompose it into semantic novelty and lexical novelty. We use the SFGram dataset \cite{sfgram}, which contains 1003 classic science fiction novels, as the reference corpus representing traditional science-fiction writing. Our framework provides a systematic and quantifiable multi-dimensional evaluation of creativity, defined as follows:

\subsubsection{Absolute Diversity}
We quantify the intrinsic diversity of generated text using token-level surprisal, i.e., the negative log-probability of each generated token under the model distribution \cite{hale-2001-probabilistic, DEMBERG2008193}. For a generated sequence of length \(L\), the average surprisal is computed as:
\begin{equation}
   S_{\text{avg}} = -\frac{1}{L} \sum_{j=1}^{L} \log p(x_j \mid x_{<j}),
\end{equation}
where \(x_j\) denotes the generated token at position \(j\), and \(p(x_j \mid x_{<j})\) is the model-assigned probability of that token given the preceding context. Concretely, we obtain next-token probabilities from the model outputs at each position and compute surprisal for the realized token \(x_j\). We normalize by \(L\) to mitigate the influence of sequence length.

Compared to entropy, which summarizes uncertainty over next-token candidates, surprisal focuses on the information content of the tokens the model actually generates, and thus provides a practical intrinsic signal for our evaluation.

\subsubsection{Lexical Divergence}
We measure lexical divergence at the unigram word level using Kullback-Leibler (KL) divergence \cite{kullback1951information}:
\begin{equation}
   D_{\text{KL}}(p \| q) = \sum_{x \in \mathcal{X}} p(x) \log \frac{p(x)}{q(x)} .
\end{equation}
Here, \(q(x)\) denotes the unigram word distribution estimated from the SFGram corpus, and \(p(x)\) denotes the unigram word distribution of the generated text. We use \(D_{\text{KL}}(p\|q)\) to quantify how the generated lexical distribution departs from the reference; larger values indicate greater lexical deviation and serve as a proxy for lexical novelty. We apply additive smoothing to avoid zero probabilities.

\subsubsection{Semantic Divergence}
\paragraph{Nearest-neighbor semantic similarity}
We measure semantic novelty by embedding generated sentences and the SFGram reference corpus, and computing cosine similarity in the embedding space. We embed SFGram using the \texttt{all-mpnet-base-v2} Sentence Transformers model \cite{reimers2019sentence}. Since the encoder has a 512-token input limit, we split SFGram into chunks of approximately 250 words and embed each chunk. For a generated sentence embedding \(\mathbf{u}\), we compute cosine similarity \cite{salton1983modern} to every reference chunk embedding \(\mathbf{v}\):
\begin{equation}
   \text{Cosine Similarity} = \frac{\mathbf{u} \cdot \mathbf{v}}{\|\mathbf{u}\| \|\mathbf{v}\|}.
\end{equation}
We take the maximum similarity over reference chunks as nearest-neighbor overlap and report semantic novelty as \(1 - \max \text{Cosine Similarity}\), so larger values indicate higher novelty.

\paragraph{Embedding Volume Gain}
In addition to nearest-neighbor semantic similarity, we measure how broadly a generated story spreads in the embedding space relative to the reference corpus. While nearest-neighbor similarity captures local overlap with the corpus, the volume gain provides a distribution-level view of semantic spread. Let \(\Sigma_{\text{ref}} = \Sigma(E_{\text{ref}})\) denote the covariance of SFGram chunk embeddings, and let \(\Sigma_{\text{ref}\cup\text{story}} = \Sigma(E_{\text{ref}} \cup E_{\text{story}})\) denote the covariance after adding the story chunk embeddings. We summarize multivariate scatter using the log-determinant of the covariance, \(\log\det(\Sigma)\), i.e., the log of the generalized variance \cite{wilks1932generalizations, rencher1998multivariate}; geometrically, \(\det(\Sigma)\) is proportional to the squared volume of the covariance ellipsoid in embedding space. We define the embedding volume gain as:
\begin{equation}
\Delta_{\text{vol}}
=
\log\det(\Sigma_{\text{ref}\cup\text{story}})
-
\log\det(\Sigma_{\text{ref}}).
\end{equation}
Larger \(\Delta_{\text{vol}}\) indicates that adding the story expands the embedding-space coverage beyond the reference corpus. For numerical stability, we compute \(\log\det(\Sigma + \epsilon I)\) with a small \(\epsilon\).

Our rule-based metrics primarily capture diversity and novelty relative to the reference corpus, rather than quality dimensions such as coherence or readability. Hence, larger deviation or dispersion may sometimes reflect increased randomness rather than better creative writing, so we interpret these metrics alongside LLM-as-a-judge scores for complementary quality-aware assessment.

\begin{table*}[ht]
    \centering
    \scriptsize
    \renewcommand{\arraystretch}{1.1}
    \setlength{\tabcolsep}{4pt}
    \begin{tabular}{lccccc|cccc}
    \toprule
    & \multicolumn{5}{c|}{\textbf{LLM-as-a-Judge Evaluation}} 
    & \multicolumn{4}{c}{\textbf{Rule-Based Evaluation}} \\
    \cmidrule(lr){2-6} \cmidrule(lr){7-10}
    \textbf{Framework} 
    & \textbf{Concepts} 
    & \textbf{Logic} 
    & \textbf{Characters} 
    & \textbf{World-Building} 
    & \textbf{Ethics}
    & \textbf{Surprisal} 
    & \textbf{KL Div.} 
    & \textbf{1-Cos Sim.} 
    & \textbf{Volume Gain} \\
    \midrule
    Single Agent           
        & 3.62$\pm$1.14 & 3.62$\pm$1.13 & 3.41$\pm$1.11 & 3.63$\pm$1.11 & 3.40$\pm$1.14
        & 0.476$\pm$0.271 & 2.233$\pm$0.012 & 0.345$\pm$0.002 & 0.0138$\pm$0.0002 \\
    LLM Teacher                
        & 3.71$\pm$0.87 & 3.73$\pm$0.87 & 3.61$\pm$0.88 & 3.73$\pm$0.87 & 3.43$\pm$0.92
        & 0.509$\pm$0.260 & 2.356$\pm$0.012 & 0.393$\pm$0.002 & 0.0160$\pm$0.0002 \\
    LLM Debate                 
        & 3.76$\pm$0.22 & 3.76$\pm$0.21 & 3.74$\pm$0.24 & 3.65$\pm$0.29 & 3.80$\pm$0.37
        & 0.540$\pm$0.456 & 2.501$\pm$0.032 & 0.424$\pm$0.006 & 0.0175$\pm$0.0003 \\
    LLM Discussion  
        & \underline{3.94$\pm$0.26} 
        & \underline{3.89$\pm$0.33} 
        & \underline{3.90$\pm$0.25} 
        & \underline{3.74$\pm$0.44} 
        & \underline{3.73$\pm$0.55}
        & \textbf{0.584$\pm$0.303}
        & \underline{2.597$\pm$0.037}
        & \underline{0.435$\pm$0.009}
        & \underline{0.0208$\pm$0.0004} \\
    LLM Review (Ours)          
        & \textbf{3.98$\pm$0.24} 
        & \textbf{4.00$\pm$0.22} 
        & \textbf{3.96$\pm$0.28} 
        & \textbf{4.04$\pm$0.26} 
        & \textbf{4.00$\pm$0.34}
        & \underline{0.573$\pm$0.316}
        & \textbf{2.638$\pm$0.034}
        & \textbf{0.441$\pm$0.008}
        & \textbf{0.0211$\pm$0.0004} \\
    \bottomrule
    \end{tabular}
    \caption{Comparison of LLM-as-a-judge and rule-based creativity evaluations on the LlaMA-3.2-3B model.}
    \label{tab:combined_llmjudge_rulebased}
\end{table*}

%% file: Section/4exp.tex
\section{Experiments}
\label{sec:experiments}

\subsection{Experimental Setup}
\label{sec:exp_setup}
We evaluate all frameworks on SciFi-100. For each prompt, the target output is a single story of approximately 300 words. Unless otherwise stated, multi-agent frameworks use $N{=}3$ writer agents with fixed personas (Section~\ref{sec:roleplay_setup}). We use consistent output-format constraints across all methods (story text only) to reduce off-format generations.

\paragraph{Decoding}
For the main comparisons, we use top\_p$=0.9$ and temperature$=0.9$ (default settings in our implementation). We additionally study the effect of decoding hyperparameters on the agent's performance.

\paragraph{Rounds and compute}
All multi-agent frameworks run for three iterative rounds. Each experiment is conducted on 4 x NVIDIA A100 Tensor Core GPUs (80GB).

\subsection{Compared Methods}
\label{sec:compared_methods}
We report results for:
(1) Single Agent, (2) LLM Discussion, (3) LLM Debate, (4) LLM Teacher, and (5) LLM Review (ours).

\subsection{Models}

We evaluate our framework across a diverse set of state-of-the-art instruction-tuned models, covering both open-weights and proprietary frontier families. For the Llama family, we utilize \textbf{Llama 3.2}~\cite{grattafiori2024llama3herdmodels}, specifically the \texttt{Llama-3.2-1B-Instruct} (llama 1b) and \texttt{Llama-3.2-3B-Instruct} (llama 3b) variants, to assess performance on lightweight, edge-class models. We also include the \textbf{Qwen 2.5} series~\cite{qwen2025qwen25technicalreport}, employing \texttt{Qwen2.5-1.5B-Instruct} (qwen 1.5b) and \texttt{Qwen2.5-3B-Instruct} (qwen 3b), to verify generalization across different model architectures. For the closed-source frontier baseline, we use \textbf{gpt-4o}~\cite{openai2024gpt4ocard}.

%% file: Section/5results.tex
\section{Results}
\label{sec:results}

\subsection{Creativity Evaluation Results}

We evaluate creativity from two complementary perspectives: an LLM-as-a-judge rubric assessing creativity-aware writing quality across five science-fiction dimensions, and a rule-based suite measuring intrinsic diversity (token-level surprisal) and novelty relative to a reference corpus via lexical (KL divergence) and semantic divergence (nearest-neighbor overlap and embedding-space volume gain against SFGram). Our main comparison uses LLaMA-3.2-3B as the writer model (Table~\ref{tab:combined_llmjudge_rulebased}).

\paragraph{Main comparison}
Across both evaluations, multi-agent frameworks outperform the single-agent baseline, with LLM Review consistently ranking highest. LLM-as-a-judge results show improvements across all five dimensions with lower score variability, indicating more robust creativity-aware writing quality. Rule-based metrics exhibit a consistent ordering, with LLM Review achieving the strongest novelty signals, followed by Discussion, while Teacher and Debate show more conservative lexical and semantic deviation from SFGram. The agreement between rubric-based judgments and automatic novelty metrics supports the interpretability of the rule-based evaluation for comparing frameworks.

\paragraph{Mechanism analysis}
The observed performance differences reflect how each framework structures interaction, feedback, and information flow. The consistent gains of LLM Review in LLM-as-a-judge scores highlight the role of explicit, targeted critique in improving creativity-aware writing quality: unlike Discussion and Debate, which expose agents to peers’ drafts without structured revision guidance, or Teacher, which provides centralized feedback that can steer agents toward similar revision targets, LLM Review decentralizes critique by requiring agents to deliver concrete peer-level feedback. At the same time, improvements in rule-based novelty metrics stem from controlled information exchange: while Discussion and Debate repeatedly condition agents on peers’ evolving outputs, encouraging alignment in phrasing and themes, LLM Review shares only independent critiques while preserving independent creative trajectories. Together, this design balances guidance and independence, yielding stronger and more stable lexical and semantic novelty signals than other multi-agent baselines.

\paragraph{Generalization across model families and scales}

Table~\ref{tab:combined_review_all_models} summarizes LLM Review’s performance across writer model families and scales. While using GPT-4o as the writer achieves the highest LLM-as-a-judge scores, its weaker rule-based novelty signals suggest potential self-preference when the writer and judge come from the same model family \citep{panickssery2024llm}, motivating reliance on rule-based metrics as a complementary reference. Across model scales within the same framework, rule-based novelty remains relatively stable, whereas LLM-as-a-judge scores increase with model size, indicating that interaction structure primarily shapes novelty while scaling mainly improves writing quality. Consistent with this, Table~\ref{tab:combined_small_vs_large} shows a favorable structure-scale trade-off: LLM Review with a smaller writer can outperform a larger single-agent baseline, supporting distributed peer feedback as an effective and relatively compute-efficient alternative to model scaling.

\begin{table*}[ht]
    \centering
    \scriptsize
    \renewcommand{\arraystretch}{1.1}
    \setlength{\tabcolsep}{4pt}
    \begin{tabular}{lccccc|ccc}
    \toprule
    & \multicolumn{5}{c|}{\textbf{LLM-as-a-Judge Evaluation}}
    & \multicolumn{3}{c}{\textbf{Rule-Based Evaluation}} \\
    \cmidrule(lr){2-6} \cmidrule(lr){7-9}
    \textbf{Model}
    & \textbf{Concepts}
    & \textbf{Logic}
    & \textbf{Characters}
    & \textbf{World-Building}
    & \textbf{Ethics}
    & \textbf{KL Div.}
    & \textbf{1-Cos Sim.}
    & \textbf{Volume Gain} \\
    \midrule
    gpt-4o
        & 4.03$\pm$0.09 & 4.07$\pm$0.18 & 4.01$\pm$0.08 & 4.22$\pm$0.29 & 4.3$\pm$0.38
        & 2.570$\pm$0.004 & 0.388$\pm$0.001 & 0.0068$\pm$0.0001 \\
    llama 3b
        & 3.98$\pm$0.24 & 4.00$\pm$0.22 & 3.96$\pm$0.28 & 4.04$\pm$0.26 & 4.00$\pm$0.34
        & 2.638$\pm$0.034 & 0.441$\pm$0.008 & 0.0211$\pm$0.0004 \\
    llama 1b
        & 3.85$\pm$0.47 & 3.81$\pm$0.50 & 3.60$\pm$0.54 & 3.75$\pm$0.50 & 3.65$\pm$0.64
        & 2.627$\pm$0.025 & 0.449$\pm$0.007 & 0.0213$\pm$0.0004 \\
    qwen 3b
        & 3.52$\pm$0.49 & 3.90$\pm$0.29 & 3.89$\pm$0.38 & 3.79$\pm$0.41 & 3.79$\pm$0.69
        & 2.254$\pm$0.008 & 0.386$\pm$0.001 & 0.0206$\pm$0.0002 \\
    qwen 1.5b
        & 3.27$\pm$0.73 & 3.41$\pm$0.73 & 3.30$\pm$0.69 & 3.39$\pm$0.70 & 3.47$\pm$0.87
        & 2.412$\pm$0.012 & 0.329$\pm$0.002 & 0.0196$\pm$0.0001 \\
    \bottomrule
    \end{tabular}
    \caption{LLM-as-a-judge and rule-based creativity evaluations across different model families for the LLM Review framework. We exclude surprisal when comparing different base models since it is model-dependent (defined under each model’s distribution $p(\cdot)$) and thus not directly comparable across models.}
    \label{tab:combined_review_all_models}
\end{table*}

\begin{table*}[ht]
    \centering
    \scriptsize
    \renewcommand{\arraystretch}{1.1}
    \setlength{\tabcolsep}{4pt}
    \begin{tabular}{lccccc|ccc}
    \toprule
    & \multicolumn{5}{c|}{\textbf{LLM-as-a-Judge Evaluation}} 
    & \multicolumn{3}{c}{\textbf{Rule-Based Evaluation}} \\
    \cmidrule(lr){2-6} \cmidrule(lr){7-9}
    \textbf{Framework} 
    & \textbf{Concepts} 
    & \textbf{Logic} 
    & \textbf{Characters} 
    & \textbf{World-Building} 
    & \textbf{Ethics}
    & \textbf{KL Div.} 
    & \textbf{1-Cos Sim.} 
    & \textbf{Volume Gain} \\
    \midrule
    Single Agent (qwen 3b)     
        & 3.09$\pm$0.68 
        & 3.25$\pm$0.67 
        & 3.24$\pm$0.46 
        & 3.13$\pm$0.61 
        & 3.18$\pm$0.82
        & 2.205$\pm$0.013 
        & \textbf{0.346$\pm$0.002}
        & 0.0148$\pm$0.0006 \\
    LLM Review (qwen 1.5b)  
        & \textbf{3.27$\pm$0.73} 
        & \textbf{3.41$\pm$0.73} 
        & \textbf{3.30$\pm$0.69} 
        & \textbf{3.39$\pm$0.70} 
        & \textbf{3.47$\pm$0.87}
        & \textbf{2.412$\pm$0.012} 
        & 0.329$\pm$0.002 
        & \textbf{0.0196$\pm$0.0001} \\
    \midrule
    Single Agent (llama 3b)           
        & 3.62$\pm$1.14 
        & 3.62$\pm$1.13 
        & 3.41$\pm$1.11 
        & 3.63$\pm$1.11 
        & 3.40$\pm$1.14
        & 2.233$\pm$0.012 
        & 0.345$\pm$0.002 
        & 0.0138$\pm$0.0002 \\
    LLM Review (llama 1b)
        & \textbf{3.85$\pm$0.47} 
        & \textbf{3.81$\pm$0.50} 
        & \textbf{3.60$\pm$0.54} 
        & \textbf{3.75$\pm$0.50} 
        & \textbf{3.65$\pm$0.64}
        & \textbf{2.627$\pm$0.025} 
        & \textbf{0.449$\pm$0.007} 
        & \textbf{0.0213$\pm$0.0004} \\
    \bottomrule
    \end{tabular}
    \caption{LLM-as-a-judge and rule-based creativity evaluations comparing smaller LLM Review models to larger single-agent baselines within the same model family. }
    \label{tab:combined_small_vs_large}
\end{table*}

\begin{table*}[ht]
    \centering
    \scriptsize
    \renewcommand{\arraystretch}{1.1}
    \setlength{\tabcolsep}{5pt}
    \begin{tabular}{lccccc}
    \toprule
    \textbf{Framework} & \textbf{Concepts} & \textbf{Logic} & \textbf{Characters} & \textbf{World-Building} & \textbf{Ethics} \\ 
\midrule
LLM Review (Human Scored) 
& 3.98$\pm$0.23 
& 3.99$\pm$0.20 
& 3.94$\pm$0.21 
& 4.02$\pm$0.23 
& 3.96$\pm$0.27 \\
\midrule
ICC(A,1)
& 0.62 
& 0.58 
& 0.59 
& 0.65 
& 0.61 \\
Bias      & 0.016 & 0.012 & -0.007 & 0.009 & -0.018 \\
LoA low             & -0.321 & -0.303 & -0.345 & -0.318 & -0.429 \\
LoA high            & 0.352 & 0.328 & 0.332 & 0.336 & 0.394 \\
Pearson $r$         & 0.679 & 0.610 & 0.607 & 0.689 & 0.647 \\

    \bottomrule
    \end{tabular}
    \caption{Human evaluation of LLM Review (llama 3B) using the same 0-5 Likert rubric and the same five dimensions as LLM-as-a-judge. We report mean$\pm$std of the averaged human scores and alignment between human consensus and LLM-as-a-judge via ICC(A,1), Bland-Altman bias and 95\% limits of agreement (LoA), and Pearson's $r$. Human ratings are consistently high across dimensions (mean 3.94-4.02). The judge shows moderate absolute agreement with humans (ICC(A,1)=0.58-0.65) and consistent linear association (Pearson's $r$=0.607-0.689). Bland-Altman analysis indicates negligible systematic bias (all $|\text{bias}|\le 0.018$), suggesting that LLM-as-a-judge scores are well-calibrated to human ratings and track human judgments in both level and ranking.}
    \label{tab:human}
 \end{table*}

 \begin{figure}[h]
    \centering
    \includegraphics[width=\linewidth]{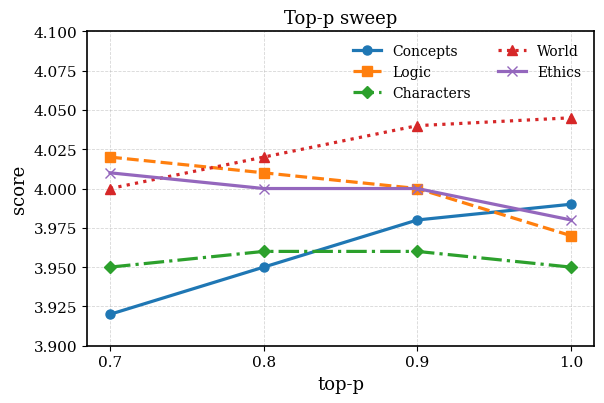}
    \caption{The average score of 5 LLM-as-a-judge evaluation aspects with different Top-p decoding methods.}
    \label{fig:topp}
\end{figure}
\begin{figure}[h]
    \centering
    \includegraphics[width=\linewidth]{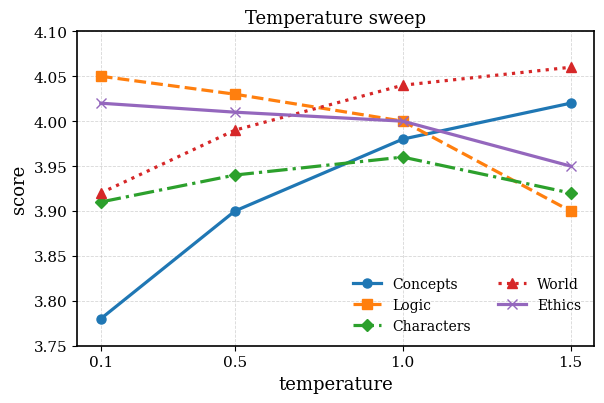}
    \caption{The average score of 5 LLM-as-a-judge evaluation aspects with different temperatures.}
    \label{fig:temp}
\end{figure}

\subsection{Human alignment}

To assess whether our LLM-as-a-judge evaluation reflects human preferences, we conduct a human rating study on outputs produced by the LLM Review framework with LlaMA-3.2-3B model. We recruited nine annotators to score each story using the same five evaluation dimensions and the same Likert rubric as in our LLM-as-a-judge setup as in Section~\ref{sec:llmasajudge}. We average the nine ratings to obtain a single human consensus score for each story and dimension, and compare it against the corresponding LLM-as-a-judge score. We quantify alignment using (i) ICC(A,1) to measure absolute agreement, (ii) Bland-Altman bias and 95\% limits of agreement (LoA) to evaluate calibration and typical per-story discrepancies, and (iii) Pearson's $r$ to capture linear association and ranking consistency between human and judge scores and the result table is shown in Table~\ref{tab:human}. The definitions of the above metrics are shown in the Section~\ref{sec:metric}.

\subsection{Ablation study}

We conduct ablation studies to understand the sensitivity of \llmreview\ to key design choices: decoding hyperparameters, number of iterative rounds, and number of participating agents. These experiments use LLaMA-3.2-3B as the writer model unless otherwise noted.

\paragraph{Decoding experiments} 
We study the effect of stochastic decoding hyperparameters on creative writing quality by sweeping top-$p$ and temperature. As shown in Figure~\ref{fig:topp}, increasing top-$p$ from 0.7 to 1.0 consistently improves scores for Scientific Concept Integration and Immersive World-Building, suggesting that allowing a broader candidate pool encourages richer idea exploration and setting construction. In contrast, Speculative Logic and Ethical/Philosophical Themes exhibit a mild downward trend, indicating a trade-off between creativity and structural coherence at higher sampling entropy. Character Depth remains relatively stable across different top-$p$ values. Figure~\ref{fig:temp} shows that increasing temperature leads to a stronger trade-off: higher temperatures substantially boost Concepts and World-Building but noticeably degrade Logic and, to a lesser extent, Ethics, while Character Depth peaks around temperature $=1.0$ before declining. Overall, these results highlight the tension between creativity and coherence in stochastic decoding and motivate our choice of a moderate top-$p$ ($\approx 0.9$) and mid-range temperature as default settings.

\begin{figure}[h]
    \centering
\includegraphics[width=\linewidth]{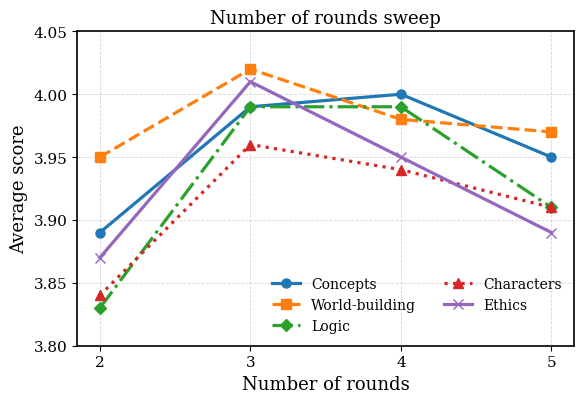}
\caption{Number of execution rounds vs the average score of 5 LLM-as-a-judge evaluation aspects.}
    \label{fig:ablation_rounds}
\end{figure}

\paragraph{Number of rounds}

We examine the effect of iterative rounds in Figure~\ref{fig:ablation_rounds}. Most dimensions achieve peak performance with three rounds of discussion. Although Concepts and Logic continue to improve slightly until the fourth round, they decline thereafter. We adopt $R=3$ rounds as our default setting to balance performance and efficiency.

\begin{figure}[H]
    \centering
    \includegraphics[width=\linewidth]{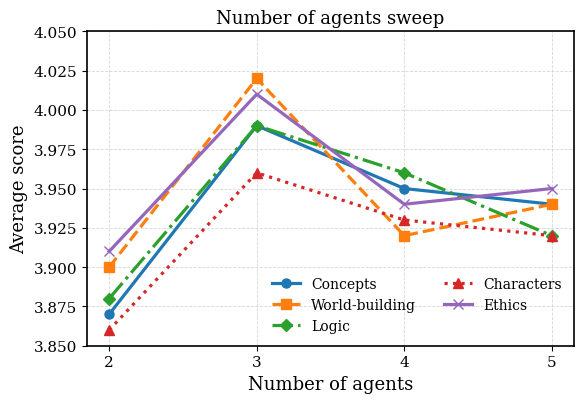}
    \caption{Number of agents vs the average score of 5 LLM-as-a-judge evaluation aspects.}
    \label{fig:ablation_agents}
\end{figure}

\paragraph{Number of agents} 

Figure~\ref{fig:ablation_agents} shows that with three-round discussions, using three agents yields the best overall performance, which declines with further additions. Unlike increasing rounds, which selectively benefits certain dimensions, adding more agents uniformly degrades all metrics, likely due to feedback dilution. We use $N=3$ agents for further experiments.

%% file: Section/6dis.tex
\section{Conclusion}
\label{sec:dis}

Our results challenge the assumption that more interaction yields better outcomes in multi-agent LLM systems. While convergence benefits tasks with verifiable ground truth, creativity requires divergence. LLM Review succeeds by disentangling feedback (targeted critique) from exposure (observing peers' outputs), where agents receive peer critique but never see how others revise. This asymmetry preserves independent creative trajectories while still benefiting from external feedback. A key finding is that smaller models using our framework outperform larger single-agent models, suggesting interaction structure may be a more compute-efficient lever than model scaling for creative tasks.

%% file: Section/7appendix.tex
\newcommand{\sciintaspectt}[1]{\colorbox{sciintcol}{#1}}
\newcommand{\speclogaspectt}[1]{\colorbox{speclogcol}{#1}}
\newcommand{\chardepaspectt}[1]{\colorbox{chardepcol}{#1}}
\newcommand{\worldaspectt}[1]{\colorbox{worldcol}{#1}}
\newcommand{\ethicsaspectt}[1]{\colorbox{ethicscol}{#1}}

\begin{table*}[!ht]
\begin{tcolorbox}[title=LLM-as-a-Judge evaluation prompt for each aspect,
  colback=white, colframe=black!60]

\textbf{System:} You are an expert evaluator for creative science fiction writing.\medskip\\
\textbf{User:}\\
You are an expert evaluator for creative science fiction writing. Evaluate the STORY independently on each of the following aspects grounded in science fiction writing literature:\medskip\\

\sciintaspect{
Scientific Concept Integration\\
- How well futuristic/scientific ideas are introduced, explained as needed, and woven into plot, setting, and character actions (not just name-dropped).\\
- Look for clarity, relevance, and meaningful impact on the story’s events.
}\medskip\\

\speclogaspect{
Speculative Logic\\
- The internal consistency and plausibility of the speculative elements given the story’s established rules.\\
- Cause-and-effect coherence, logical constraints, and avoidance of convenient contradictions.
}\medskip\\

\chardepaspect{
Character Depth\\
- Distinct, believable characters with motivations, agency, and emotional/psychological complexity.\\
- Growth, conflict, or meaningful choices that feel earned.
}\medskip\\

\worldaspect{
Immersive World-Building\\
- A vivid, coherent story world with concrete details (culture, environment, technology, institutions) that support immersion.\\
- World details should serve the narrative rather than overwhelm it.
}\medskip\\

\ethicsaspect{
Ethical and Philosophical Themes\\
- Presence and depth of ethical, societal, or philosophical questions typical of strong science fiction.\\
- Nuance, originality, and integration into narrative stakes (not purely didactic).
}\medskip\\

Scoring scale (0-5) for each of the above aspects:\\
- 0: Strong indicators of potential plagiarism OR extremely low-quality/derivative writing with minimal original content.\\
- 1: Very weak; major flaws, thin originality, little coherence or craft.\\
- 2: Weak-to-fair; some promising ideas but underdeveloped or inconsistent execution.\\
- 3: Competent; clear strengths with noticeable but non-fatal weaknesses.\\
- 4: Strong; well-executed, creative, and cohesive with only minor issues.\\
- 5: Exceptional; highly creative, polished, and deeply integrated execution across the aspect.\medskip\\

STORY:\\
\{story\}\medskip\\

SCORE:
\end{tcolorbox}
\caption{Prompt template for different LLM-as-a-Judge evaluation aspects (\sciintaspectt{Scientific Concept Integration}, \speclogaspectt{Speculative Logic}, \chardepaspectt{Character Depth}, \worldaspectt{Immersive World-Building}, \ethicsaspectt{Ethical and Philosophical Themes}) and each aspect is evaluated independently. Human annotators use the same prompt except for the system prompt part.}
\label{tab:evaluation_prompts}
\end{table*}

\section{SciFi-100 Overview}\label{apen:scifi-100}
\begin{figure}[ht]
\begin{center}
\includegraphics[width=1\linewidth]{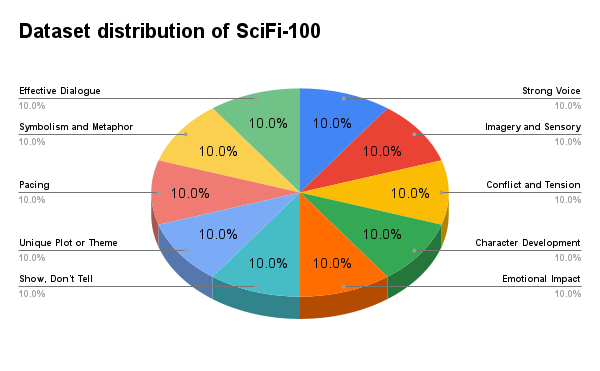}
\end{center}
\caption{Dataset distribution of SciFi-100.
}
\label{fig:scifi100}
\end{figure}

Figure \ref{fig:scifi100} shows the data distribution of our SciFi-100 and Table \ref{tab:scifi-100_prompts} shows example prompts from each aspect of creative writing.

\begin{table*}[ht]
\centering
\small
\begin{tabular}{@{}lp{12cm}@{}}
\toprule
\textbf{Aspect of Creative Writing} & \textbf{Example Prompt} \\ \midrule

\textbf{Strong Voice} & Write from the perspective of a starship mechanic on a long-haul journey who discovers a hidden compartment. Let their tone be equal parts suspicion, curiosity, and a touch of rebellious glee.
     \\ 
     \midrule
\textbf{Imagery and Sensory Details} & Capture the awe and danger as a spacecraft crew observes a neutron star up close, each sensation amplifying the thrill and peril of the sight.
     \\ 
     \midrule
\textbf{Conflict and Tension} & Show us a researcher trying to persuade a skeptical alien committee to support a high-risk, high-reward interspecies research project.
     \\ 
     \midrule
\textbf{Character Development} & Tell the tale of a researcher overcoming a deep-seated fear of spacewalks to collect crucial samples from an asteroid.
     \\ 
     \midrule
\textbf{Emotional Impact} & Convey the exhilaration of a scientist who finally finds evidence of life on another planet after decades of searching, a discovery beyond their wildest dreams.
     \\ 
     \midrule
\textbf{Show, Don’t Tell} & Capture a character’s unquenchable curiosity about alien species through their late nights, stacks of research, and endless notebook scribbles.
     \\ 
     \midrule
\textbf{Unique Plot or Theme} & Depict a futuristic lab with live-streamed alien dissections, complete with real-time public commentary and reactions.
     \\ 
     \midrule
\textbf{Pacing} & Imagine the team dynamic as anticipation builds for a groundbreaking discovery presentation, followed by a subdued, introspective moment afterward.
     \\ 
     \midrule
\textbf{Symbolism and Metaphor} & Use the image of a sealed, heavily guarded lab door to symbolize the secrecy and exclusivity surrounding a groundbreaking experiment.
     \\ 
     \midrule
\textbf{Effective Dialogue} & Imagine a heated conversation between two scientists debating the ethical implications of weaponizing a newly discovered alien technology.
    \\ 
    
    \bottomrule
\end{tabular}
\caption{Example prompts from SciFi-100 by aspects of creative writing.}
\label{tab:scifi-100_prompts}
\end{table*}

\section{LLM-as-a-Judge Evaluation Prompts}
\label{sec:appendix}

Table \ref{tab:evaluation_prompts} shows our prompts for LLM-as-a-Judge evaluation on five different criterias for scientific fiction creativity writing.

\section{Alignment Metrics}
\label{sec:metric}
For each story $i\in\{1,\dots,N\}$ and evaluation dimension $d$, let $h_{i,d}\in[0,5]$ denote the human
consensus score (averaged over annotators) and $s_{i,d}\in[0,5]$ denote the LLM-as-a-judge score.
We compute alignment \emph{separately for each dimension $d$}. For clarity, we fix a dimension $d$
and omit the subscript $d$ below, writing $h_i$ and $s_i$.

\paragraph{(1) ICC(A,1): absolute agreement (two-way random effects, single measurement)}
We treat the human consensus and the LLM judge as two ``raters'' ($k=2$) that score the same $N$
targets (stories). Define the score matrix $X\in\mathbb{R}^{N\times k}$ by
\[
x_{i1}=h_i,\quad x_{i2}=s_i .
\]
Let the row mean, column mean be:
\begin{equation}
    \bar{x}_{i\cdot}=\frac{1}{k}\sum_{j=1}^k x_{ij},\quad
\bar{x}_{\cdot j}=\frac{1}{N}\sum_{i=1}^N x_{ij},
\end{equation}
And the grand mean be:
\begin{equation}
    \bar{x}_{\cdot\cdot}=\frac{1}{Nk}\sum_{i=1}^N\sum_{j=1}^k x_{ij}
\end{equation}
The ANOVA mean squares are
\begin{equation}
    MS_R=\frac{k}{N-1}\sum_{i=1}^N(\bar{x}_{i\cdot}-\bar{x}_{\cdot\cdot})^2
\end{equation}
\begin{equation}
    MS_C=\frac{N}{k-1}\sum_{j=1}^k(\bar{x}_{\cdot j}-\bar{x}_{\cdot\cdot})^2
\end{equation}
\begin{equation}
    MS_E=\frac{\sum_{i=1}^N\sum_{j=1}^k
\left(x_{ij}-\bar{x}_{i\cdot}-\bar{x}_{\cdot j}+\bar{x}_{\cdot\cdot}\right)^2}{(N-1)(k-1)} 
\end{equation}
The intraclass correlation coefficient for absolute agreement, single measurement is
\begin{equation}
    \mathrm{ICC(A,1)} =
\frac{MS_R - MS_E}{MS_R + \frac{(N-1)k-N}{N}MS_E + \frac{k}{N}MS_C}
\end{equation}

Higher ICC indicates stronger absolute agreement ($1$ is perfect agreement); values can be close to
$0$ (weak agreement) or even negative when disagreement dominates.

\paragraph{(2) Pearson correlation: linear association and ranking consistency}
Pearson's $r$ between $\{h_i\}_{i=1}^N$ and $\{s_i\}_{i=1}^N$ is
\begin{equation}
    r =
\frac{\sum_{i=1}^N (h_i-\bar{h})(s_i-\bar{s})}
{\sqrt{\sum_{i=1}^N (h_i-\bar{h})^2}\;\sqrt{\sum_{i=1}^N (s_i-\bar{s})^2}}
\end{equation}
where $\bar{h}=\frac{1}{N}\sum_{i=1}^N h_i$ and $\bar{s}=\frac{1}{N}\sum_{i=1}^N s_i$.
This metric captures whether stories that humans score higher also tend to receive higher judge scores.

\paragraph{(3) Bland-Altman: calibration via bias and 95\% limits of agreement (LoA).}
Define the per-story difference and (optionally) the per-story mean as
\begin{equation}
    \Delta_i = s_i-h_i
\end{equation}
The \textbf{bias} (mean signed difference) is
\begin{equation}
    \mathrm{bias} = \bar{\Delta} = \frac{1}{N}\sum_{i=1}^N \Delta_i
\end{equation}
and the sample standard deviation of differences is
\begin{equation}
    SD_\Delta = \sqrt{\frac{1}{N-1}\sum_{i=1}^N (\Delta_i-\bar{\Delta})^2}
\end{equation}
Assuming the differences are approximately normally distributed, the \textbf{95\% limits of agreement}
are
\begin{equation}
    \mathrm{LoA}_{\mathrm{low,high}}=\bar{\Delta} \pm 1.96\,SD_\Delta
\end{equation}
which estimate the interval in which the judge-human discrepancy $\Delta_i$ is expected to fall for
about 95\% of stories.

\section{Potential Risks}
The proposed framework may amplify harmful or biased narratives present in underlying language models, as multi-agent critique and revision can reinforce shared assumptions rather than surface alternative viewpoints. In addition, LLM Review could be misused to automate large-scale creative content generation, contributing to content flooding and reducing the visibility of human authorship. Finally, its human-inspired design may encourage over-attribution of agency or originality to machine-generated outputs, highlighting the need for careful deployment and human oversight.

\section{The Use of Large Language Models (LLMs)}
\label{app:llm-usage}
LLM was used only to aid writing quality (proofreading and polishing grammar) and generate the SciFi-100 dataset. No ideas, claims, methods, results, or references are generated by LLMs. All content decisions and revisions are made by the authors.
\section{Human Evaluation Protocol: Participant Instructions, Recruitment,}
\label{app:human-eval-protocol}

\subsection{Instructions Given to Participants}
\label{app:participant-instructions}

\paragraph{Study overview}
You are invited to take part in a research study about evaluating short science-fiction stories.
In this task, you will read a set of short stories (approximately 300 words each) and rate them on several
quality/creativity-related dimensions. The stories you will read are machine-generated.

\medskip
\paragraph{What you will do}
For each story, you will provide \textbf{five separate ratings} (integers from 0 to 5) according to the criteria below: (This part is same as Table~\ref{tab:evaluation_prompts} so we skipped here. )

\medskip
\paragraph{Important guidelines}
\begin{itemize}
  \item Provide your \textbf{independent judgment}. There are no right or wrong answers.
  \item Use the \textbf{full 0-5 scale} when appropriate.
  \item Rate the story \textbf{as written}; do not assume missing details unless implied by the text.
  \item Do \textbf{not} spend time proofreading grammar; focus on the five criteria above.
  \item If you are unsure between two scores, choose the one that best matches the rubric definitions.
\end{itemize}

\medskip
\paragraph{Risks and sensitive content notice}
This is a minimal-risk study. However, because the content is science fiction, some stories may include
fictional depictions of conflict, danger, or other potentially sensitive themes.
If you feel uncomfortable at any time, you may stop immediately or skip a story without penalty.

\medskip
\paragraph{Privacy and data handling}
We record only your story ratings for analysis. We do not ask you to provide personal identifying information
as part of the ratings task, except what may be needed to administer compensation (if applicable).
We report results only in aggregate.

\subsection{Recruitment}
\label{app:recruitment-payment}

We recruited \textbf{nine student annotators} to rate machine-generated stories produced for prompts from SciFi-100.
Participants were recruited via university mailing.
Inclusion criteria were: (i) age 18 or older, (ii) proficient in English reading comprehension, and (iii) willingness to read and rate short science-fiction stories.

%% file: custom.bib
@article{chan2023chatevalbetterllmbasedevaluators,
  title={Chateval: Towards better llm-based evaluators through multi-agent debate},
  author={Chan, Chi-Min and Chen, Weize and Su, Yusheng and Yu, Jianxuan and Xue, Wei and Zhang, Shanghang and Fu, Jie and Liu, Zhiyuan},
  journal={arXiv preprint arXiv:2308.07201},
  year={2023}
}

@book{rencher1998multivariate,
  title={Multivariate statistical inference and applications},
  author={Rencher, Alvin C},
  volume={635},
  year={1998},
  publisher={Wiley New York}
}

@inproceedings{hale-2001-probabilistic,
  title={A probabilistic Earley parser as a psycholinguistic model},
  author={Hale, John},
  booktitle={Second meeting of the north american chapter of the association for computational linguistics},
  year={2001}
}

@article{DEMBERG2008193,
  title={Data from eye-tracking corpora as evidence for theories of syntactic processing complexity},
  author={Demberg, Vera and Keller, Frank},
  journal={Cognition},
  volume={109},
  number={2},
  pages={193--210},
  year={2008},
  publisher={Elsevier}
}

@article{wilks1932generalizations,
  title={Certain generalizations in the analysis of variance},
  author={Wilks, Samuel S},
  journal={Biometrika},
  volume={24},
  number={3/4},
  pages={471--494},
  year={1932},
  publisher={JSTOR}
}

@article{tseng2024talespersonallmssurvey,
  title={Two tales of persona in llms: A survey of role-playing and personalization},
  author={Tseng, Yu-Min and Huang, Yu-Chao and Hsiao, Teng-Yun and Chen, Wei-Lin and Huang, Chao-Wei and Meng, Yu and Chen, Yun-Nung},
  journal={arXiv preprint arXiv:2406.01171},
  year={2024}
}

@article{lu2024llmdiscussionenhancingcreativity,
  title={LLM discussion: Enhancing the creativity of large language models via discussion framework and role-play},
  author={Lu, Li-Chun and Chen, Shou-Jen and Pai, Tsung-Min and Yu, Chan-Hung and Lee, Hung-yi and Sun, Shao-Hua},
  journal={arXiv preprint arXiv:2405.06373},
  year={2024}
}

@inproceedings{zhang2020tradingdiversityqualitynatural,
  title={Trading off diversity and quality in natural language generation},
  author={Zhang, Hugh and Duckworth, Daniel and Ippolito, Daphne and Neelakantan, Arvind},
  booktitle={Proceedings of the workshop on Human Evaluation of NLP Systems (HumEval)},
  pages={25--33},
  year={2021}
}

@article{peeperkorn2024temperaturecreativityparameterlarge,
  title={Is temperature the creativity parameter of large language models?},
  author={Peeperkorn, Max and Kouwenhoven, Tom and Brown, Dan and Jordanous, Anna},
  journal={arXiv preprint arXiv:2405.00492},
  year={2024}
}

@article{gómezrodríguez2023confederacymodelscomprehensiveevaluation,
  title={A confederacy of models: A comprehensive evaluation of LLMs on creative writing},
  author={G{\'o}mez-Rodr{\'\i}guez, Carlos and Williams, Paul},
  journal={arXiv preprint arXiv:2310.08433},
  year={2023}
}

@article{doi:10.1080/10400419.2024.2326343,
  title={Automatic scoring of metaphor creativity with large language models},
  author={DiStefano, Paul V and Patterson, John D and Beaty, Roger E},
  journal={Creativity Research Journal},
  volume={37},
  number={4},
  pages={555--569},
  year={2025},
  publisher={Taylor \& Francis}
}

@article{stevenson2022puttinggpt3screativityalternative,
  title={Putting GPT-3's creativity to the (alternative uses) test},
  author={Stevenson, Claire and Smal, Iris and Baas, Matthijs and Grasman, Raoul and van der Maas, Han},
  journal={arXiv preprint arXiv:2206.08932},
  year={2022}
}

@inproceedings{colton2008creativity,
  title={Creativity Versus the Perception of Creativity in Computational Systems.},
  author={Colton, Simon},
  booktitle={AAAI spring symposium: creative intelligent systems},
  volume={8},
  pages={7},
  year={2008},
  organization={Palo Alto, CA}
}

@article{d2021characterises,
  title={What characterises creativity in narrative writing, and how do we assess it? Research findings from a systematic literature search},
  author={D'Souza, Richard},
  journal={Thinking skills and creativity},
  volume={42},
  pages={100949},
  year={2021},
  publisher={Elsevier}
}

@inproceedings{yao2022react,
  title={React: Synergizing reasoning and acting in language models},
  author={Yao, Shunyu and Zhao, Jeffrey and Yu, Dian and Du, Nan and Shafran, Izhak and Narasimhan, Karthik R and Cao, Yuan},
  booktitle={The eleventh international conference on learning representations},
  year={2022}
}

@inproceedings{chen2023reconcile,
  title={Reconcile: Round-table conference improves reasoning via consensus among diverse llms},
  author={Chen, Justin and Saha, Swarnadeep and Bansal, Mohit},
  booktitle={Proceedings of the 62nd Annual Meeting of the Association for Computational Linguistics (Volume 1: Long Papers)},
  pages={7066--7085},
  year={2024}
}

@inproceedings{hong2024metagpt,
  title={MetaGPT: Meta programming for a multi-agent collaborative framework},
  author={Hong, Sirui and Zhuge, Mingchen and Chen, Jonathan and Zheng, Xiawu and Cheng, Yuheng and Wang, Jinlin and Zhang, Ceyao and Wang, Zili and Yau, Steven Ka Shing and Lin, Zijuan and others},
  booktitle={The Twelfth International Conference on Learning Representations},
  year={2023}
}

@inproceedings{qian-etal-2024-chatdev,
  title={Chatdev: Communicative agents for software development},
  author={Qian, Chen and Liu, Wei and Liu, Hongzhang and Chen, Nuo and Dang, Yufan and Li, Jiahao and Yang, Cheng and Chen, Weize and Su, Yusheng and Cong, Xin and others},
  booktitle={Proceedings of the 62nd Annual Meeting of the Association for Computational Linguistics (Volume 1: Long Papers)},
  pages={15174--15186},
  year={2024}
}

@article{lappin2024assessing,
  title={Assessing the strengths and weaknesses of large language models},
  author={Lappin, Shalom},
  journal={Journal of Logic, Language and Information},
  volume={33},
  number={1},
  pages={9--20},
  year={2024},
  publisher={Springer}
}

@article{jin2024comprehensive,
  title={A comprehensive survey on process-oriented automatic text summarization with exploration of llm-based methods},
  author={Zhang, Yang and Jin, Hanlei and Meng, Dan and Wang, Jun and Tan, Jinghua},
  journal={arXiv preprint arXiv:2403.02901},
  year={2024}
}

@article{yi2024survey,
  title={A survey on recent advances in llm-based multi-turn dialogue systems},
  author={Yi, Zihao and Ouyang, Jiarui and Xu, Zhe and Liu, Yuwen and Liao, Tianhao and Luo, Haohao and Shen, Ying},
  journal={ACM Computing Surveys},
  year={2024},
  publisher={ACM New York, NY}
}

@article{mohammadi2024creativity,
  title={Creativity has left the chat: The price of debiasing language models},
  author={Mohammadi, Behnam},
  journal={arXiv preprint arXiv:2406.05587},
  year={2024}
}

@inproceedings{chakrabarty2024art,
  title={Art or artifice? large language models and the false promise of creativity},
  author={Chakrabarty, Tuhin and Laban, Philippe and Agarwal, Divyansh and Muresan, Smaranda and Wu, Chien-Sheng},
  booktitle={Proceedings of the 2024 CHI Conference on Human Factors in Computing Systems},
  pages={1--34},
  year={2024}
}

@article{zheng2023judging,
  title={Judging llm-as-a-judge with mt-bench and chatbot arena},
  author={Zheng, Lianmin and Chiang, Wei-Lin and Sheng, Ying and Zhuang, Siyuan and Wu, Zhanghao and Zhuang, Yonghao and Lin, Zi and Li, Zhuohan and Li, Dacheng and Xing, Eric and others},
  journal={Advances in neural information processing systems},
  volume={36},
  pages={46595--46623},
  year={2023}
}

@inproceedings{feng2024sampleefficienthumanevaluationlarge,
  title={Sample-efficient human evaluation of large language models via maximum discrepancy competition},
  author={Feng, Kehua and Ding, Keyan and Hongzhi, Tan and Ma, Kede and Wang, Zhihua and Guo, Shuangquan and Yuzhou, Cheng and Sun, Ge and Zheng, Guozhou and Zhang, Qiang and others},
  booktitle={Proceedings of the 63rd Annual Meeting of the Association for Computational Linguistics (Volume 1: Long Papers)},
  pages={10913--10947},
  year={2025}
}

@inproceedings{yang2024largelanguagemodelsoptimizers,
  title={Large language models as optimizers},
  author={Yang, Chengrun and Wang, Xuezhi and Lu, Yifeng and Liu, Hanxiao and Le, Quoc V and Zhou, Denny and Chen, Xinyun},
  booktitle={The Twelfth International Conference on Learning Representations},
  year={2023}
}

@inproceedings{summers2023brainstorm,
  title={Brainstorm, then select: a generative language model improves its creativity score},
  author={Summers-Stay, Douglas and Voss, Clare R and Lukin, Stephanie M},
  booktitle={The AAAI-23 Workshop on Creative AI Across Modalities},
  year={2023}
}

@inproceedings{ghazvininejad2017hafez,
  title={Hafez: an interactive poetry generation system},
  author={Ghazvininejad, Marjan and Shi, Xing and Priyadarshi, Jay and Knight, Kevin},
  booktitle={Proceedings of ACL 2017, System Demonstrations},
  pages={43--48},
  year={2017}
}

@article{keskar2019ctrl,
  title={Ctrl: A conditional transformer language model for controllable generation},
  author={Keskar, Nitish Shirish and McCann, Bryan and Varshney, Lav R and Xiong, Caiming and Socher, Richard},
  journal={arXiv preprint arXiv:1909.05858},
  year={2019}
}

@inproceedings{Du2023ImprovingFA,
  title={Improving factuality and reasoning in language models through multiagent debate},
  author={Du, Yilun and Li, Shuang and Torralba, Antonio and Tenenbaum, Joshua B and Mordatch, Igor},
  booktitle={Forty-first International Conference on Machine Learning},
  year={2023}
}

@misc{sfgram,
  author = {Schaetti, Nils},
  title = {SFGram: a dataset containing thousands of scienc-fiction books and novels},
  year = {2018},
  publisher = {GitHub},
  journal = {GitHub repository},
  howpublished = {\url{https://github.com/nschaetti/EchoTorch}},
}

@article{design_thinking,
  title={David Kelley: From design to design thinking at Stanford and IDEO},
  author={Camacho, Maria},
  journal={She Ji: The Journal of Design, Economics, and Innovation},
  volume={2},
  number={1},
  pages={88--101},
  year={2016},
  publisher={Elsevier}
}

@article{kullback1951information,
  title={On information and sufficiency},
  author={Kullback, Solomon and Leibler, Richard A},
  journal={The annals of mathematical statistics},
  volume={22},
  number={1},
  pages={79--86},
  year={1951},
  publisher={JSTOR}
}

@inproceedings{salton1983modern,
  title={Introduction to Modern Information Retrieval},
  author={Gerard Salton and Michael McGill},
  year={1983},
  url={https://api.semanticscholar.org/CorpusID:43685115}
}

@article{reimers2019sentence,
  title={Sentence-bert: Sentence embeddings using siamese bert-networks},
  author={Reimers, Nils and Gurevych, Iryna},
  journal={arXiv preprint arXiv:1908.10084},
  year={2019}
}

@inproceedings{wei2025ignitingcreativewritingsmall,
  title={Igniting Creative Writing in Small Language Models: LLM-as-a-Judge versus Multi-Agent Refined Rewards},
  author={Wei, Xiaolong and Lu, Bo and Zhang, Xingyu and Zhao, Zhejun and Shen, Dongdong and Xia, Long and Yin, Dawei},
  booktitle={Proceedings of the 2025 Conference on Empirical Methods in Natural Language Processing},
  pages={17171--17197},
  year={2025}
}

@article{chung2025modifyinglargelanguagemodel,
  title={Modifying Large Language Model Post-Training for Diverse Creative Writing},
  author={Chung, John Joon Young and Padmakumar, Vishakh and Roemmele, Melissa and Sun, Yuqian and Kreminski, Max},
  journal={arXiv preprint arXiv:2503.17126},
  year={2025}
}

@article{lagzian2025multinoveltyimprovediversitynovelty,
  title={Multi-Novelty: Improve the Diversity and Novelty of Contents Generated by Large Language Models via inference-time Multi-Views Brainstorming},
  author={Lagzian, Arash and Anumasa, Srinivas and Liu, Dianbo},
  journal={arXiv preprint arXiv:2502.12700},
  year={2025}
}

@article{qian2025scaling_macnet,
  title={Scaling large language model-based multi-agent collaboration},
  author={Qian, Chen and Xie, Zihao and Wang, Yifei and Liu, Wei and Zhu, Kunlun and Xia, Hanchen and Dang, Yufan and Du, Zhuoyun and Chen, Weize and Yang, Cheng and others},
  journal={arXiv preprint arXiv:2406.07155},
  year={2024}
}

@article{zhang2025cut,
  title={Cut the crap: An economical communication pipeline for llm-based multi-agent systems},
  author={Zhang, Guibin and Yue, Yanwei and Li, Zhixun and Yun, Sukwon and Wan, Guancheng and Wang, Kun and Cheng, Dawei and Yu, Jeffrey Xu and Chen, Tianlong},
  journal={arXiv preprint arXiv:2410.02506},
  year={2024}
}

@article{li2025agent_oriented_planning,
  title={Agent-oriented planning in multi-agent systems},
  author={Li, Ao and Xie, Yuexiang and Li, Songze and Tsung, Fugee and Ding, Bolin and Li, Yaliang},
  journal={arXiv preprint arXiv:2410.02189},
  year={2024}
}

@article{dang2025evolving_orchestration,
  title={Multi-Agent Collaboration via Evolving Orchestration},
  author={Dang, Yufan and Qian, Chen and Luo, Xueheng and Fan, Jingru and Xie, Zihao and Shi, Ruijie and Chen, Weize and Yang, Cheng and Che, Xiaoyin and Tian, Ye and others},
  journal={arXiv preprint arXiv:2505.19591},
  year={2025}
}

@article{ye2025mas_gpt,
  title={MAS-GPT: Training LLMs to build LLM-based multi-agent systems},
  author={Ye, Rui and Tang, Shuo and Ge, Rui and Du, Yaxin and Yin, Zhenfei and Chen, Siheng and Shao, Jing},
  journal={arXiv preprint arXiv:2503.03686},
  year={2025}
}

@inproceedings{wang-etal-2025-megaagent,
  title={MegaAgent: A large-scale autonomous LLM-based multi-agent system without predefined SOPs},
  author={Wang, Qian and Wang, Tianyu and Tang, Zhenheng and Li, Qinbin and Chen, Nuo and Liang, Jingsheng and He, Bingsheng},
  booktitle={Findings of the Association for Computational Linguistics: ACL 2025},
  pages={4998--5036},
  year={2025}
}

@article{tran2025multiagentcollaborationmechanismssurvey,
  title={Multi-agent collaboration mechanisms: A survey of llms},
  author={Tran, Khanh-Tung and Dao, Dung and Nguyen, Minh-Duong and Pham, Quoc-Viet and O'Sullivan, Barry and Nguyen, Hoang D},
  journal={arXiv preprint arXiv:2501.06322},
  year={2025}
}

@article{li2025automated,
  title={Automated Creativity Evaluation for Large Language Models: A Reference-Based Approach},
  author={Li, Ruizhe and Zhu, Chiwei and Xu, Benfeng and Wang, Xiaorui and Mao, Zhendong},
  journal={arXiv preprint arXiv:2504.15784},
  year={2025}
}

@article{potraghloo2025tophdecodingadaptingcreativity,
  title={Top-h decoding: Adapting the creativity and coherence with bounded entropy in text generation},
  author={Potraghloo, Erfan Baghaei and Azizi, Seyedarmin and Kundu, Souvik and Pedram, Massoud},
  journal={arXiv preprint arXiv:2509.02510},
  year={2025}
}

@inproceedings{matan2025comprehensive,
  title={A Comprehensive Review of Supervised Fine-Tuning for Large Language Models in Creative Applications and Content Moderation},
  author={Matan, P and Velvizhy, P},
  booktitle={2025 International Conference on Inventive Computation Technologies (ICICT)},
  pages={1294--1299},
  year={2025},
  organization={IEEE}
}

@book{genette1980narrative,
  title={Narrative discourse: An essay in method},
  author={Genette, G{\'e}rard},
  volume={3},
  year={1980},
  publisher={Cornell University Press}
}

@book{pound2013abc,
  title={ABC of Reading},
  author={Pound, Ezra},
  year={2013},
  publisher={New Directions Publishing}
}

@book{freytag1895technique,
  title={Technique of the drama: An exposition of dramatic composition and art},
  author={Freytag, Gustav},
  year={1895},
  publisher={S. Griggs}
}

@book{forster1927aspects,
  title={Aspects of the Novel},
  author={Forster, Edward Morgan},
  year={1927},
  publisher={Harcourt, Brace}
}

@book{rosenblatt1994reader,
  title={The reader, the text, the poem: The transactional theory of the literary work},
  author={Rosenblatt, Louise M},
  year={1994},
  publisher={SIU Press}
}

@book{burroway2022writing,
  title={Writing fiction: A guide to narrative craft},
  author={Burroway, Janet and Stuckey-French, Elizabeth and Stuckey-French, Ned},
  year={2022},
  publisher={University of Chicago Press}
}

@article{csikszentmihalyi1990domain,
  title={The domain of creativity.},
  author={Csikszentmihalyi, Mihaly},
  year={1990},
  publisher={Sage Publications, Inc}
}

@book{chatman1978story,
  title={Story and discourse: Narrative structure in fiction and film},
  author={Chatman, Seymour Benjamin and Chatman, Seymour},
  year={1978},
  publisher={Cornell university press}
}

@book{ricoeur2004rule,
  title={The rule of metaphor: The creation of meaning in language},
  author={Ricoeur, Paul},
  year={2004},
  publisher={Routledge}
}

@book{tannen2005conversational,
  title={Conversational style: Analyzing talk among friends},
  author={Tannen, Deborah},
  year={2005},
  publisher={Oxford University Press}
}

@article{qwen2025qwen25technicalreport,
  title={Qwen3 technical report},
  author={Yang, An and Li, Anfeng and Yang, Baosong and Zhang, Beichen and Hui, Binyuan and Zheng, Bo and Yu, Bowen and Gao, Chang and Huang, Chengen and Lv, Chenxu and others},
  journal={arXiv preprint arXiv:2505.09388},
  year={2025}
}

@article{grattafiori2024llama3herdmodels,
  title={The llama 3 herd of models},
  author={Grattafiori, Aaron and Dubey, Abhimanyu and Jauhri, Abhinav and Pandey, Abhinav and Kadian, Abhishek and Al-Dahle, Ahmad and Letman, Aiesha and Mathur, Akhil and Schelten, Alan and Vaughan, Alex and others},
  journal={arXiv preprint arXiv:2407.21783},
  year={2024}
}

@article{openai2024gpt4ocard,
  title={Gpt-4o system card},
  author={Hurst, Aaron and Lerer, Adam and Goucher, Adam P and Perelman, Adam and Ramesh, Aditya and Clark, Aidan and Ostrow, AJ and Welihinda, Akila and Hayes, Alan and Radford, Alec and others},
  journal={arXiv preprint arXiv:2410.21276},
  year={2024}
}

@article{canavan2016metamorphoses,
  title={Metamorphoses of science fiction},
  author={Canavan, Gerry and Suvin, Darko},
  year={2016},
  publisher={Peter Lang Publishing}
}

@book{le2015steering,
  title={Steering the craft: A twenty-first century guide to sailing the sea of story},
  author={Le Guin, Ursula K},
  year={2015},
  publisher={Houghton Mifflin Harcourt}
}

@article{nussbaum1988love,
  title={Love's Knowledge},
  author={Nussbaum, Martha},
  journal={Perspectives on Self-Deception},
  pages={488--514},
  year={1988},
  publisher={University of California Press Berkeley}
}

@article{panickssery2024llm,
  title={Llm evaluators recognize and favor their own generations},
  author={Panickssery, Arjun and Bowman, Samuel and Feng, Shi},
  journal={Advances in Neural Information Processing Systems},
  volume={37},
  pages={68772--68802},
  year={2024}
}

@incollection{paulus2003group,
    author = {Nijstad, Bernard A. and Paulus, Paul B.},
    isbn = {9780195147308},
    title = {Group Creativity: Common Themes and Future Directions},
    booktitle = {Group Creativity: Innovation through Collaboration},
    publisher = {Oxford University Press},
    year = {2003},
    month = {09},
    doi = {10.1093/acprof:oso/9780195147308.003.0015},
    url = {https://doi.org/10.1093/acprof:oso/9780195147308.003.0015},
    eprint = {https://academic.oup.com/book/0/chapter/196541000/chapter-ag-pdf/44610201/book_27143_section_196541000.ag.pdf},
}

@article{diehl1987productivity,
  title={Productivity loss in brainstorming groups: Toward the solution of a riddle},
  author={Diehl, Michael and Stroebe, Wolfgang},
  journal={Journal of Personality and Social Psychology},
  volume={53},
  number={3},
  pages={497--509},
  year={1987}
}

@inproceedings{anderson2024homogenization,
  title={Homogenization effects of large language models on human creative ideation},
  author={Anderson, Brenda R and Shah, Jasneet H and Kreminski, Max},
  booktitle={Proceedings of the 16th Conference on Creativity \& Cognition},
  pages={413--425},
  year={2024}
}

@article{larey1999group,
  title={Group preference and convergent tendencies in small groups: A content analysis of group brainstorming performance},
  author={Larey, Timothy S and Paulus, Paul B},
  journal={Creativity Research Journal},
  volume={12},
  number={3},
  pages={175--184},
  year={1999}
}

@article{gillebaart2013unraveling,
  title={Unraveling effects of novelty on creativity},
  author={Gillebaart, Marleen and F{\"o}rster, Jens and Rotteveel, Mark and Jehle, Astrid CM},
  journal={Creativity Research Journal},
  volume={25},
  number={3},
  pages={280--285},
  year={2013},
  publisher={Taylor \& Francis}
}

@article{li2026grading,
  title={Grading Scale Impact on LLM-as-a-Judge: Human-LLM Alignment Is Highest on 0-5 Grading Scale},
  author={Li, Weiyue and Zhao, Minda and Dong, Weixuan and Cai, Jiahui and Wei, Yuze and Pocress, Michael and Li, Yi and Yuan, Wanyan and Wang, Xiaoyue and Hou, Ruoyu and others},
  journal={arXiv preprint arXiv:2601.03444},
  year={2026}
}
